\def\year{2022}\relax
\documentclass[letterpaper]{article} 
\usepackage{aaai22}  
\usepackage{times}  
\usepackage{helvet}  
\usepackage{courier}  
\usepackage[hyphens]{url}  
\usepackage{graphicx} 
\usepackage{natbib}  
\usepackage{caption} 
\DeclareCaptionStyle{ruled}{labelfont=normalfont,labelsep=colon,strut=off} 
\usepackage{algorithm}
\usepackage{algorithmic}
\usepackage{amssymb}
\usepackage{amsmath}
\usepackage{multirow}
\usepackage{booktabs}

\usepackage{newfloat}
\usepackage{listings}
\floatstyle{ruled}
\newfloat{listing}{tb}{lst}{}
\floatname{listing}{Listing}
\title{ContrastNet: A Contrastive Learning Framework for Few-Shot Text Classification}
\author{
    Junfan Chen\textsuperscript{1},
    Richong Zhang\textsuperscript{1}\thanks{Corresponding author: zhangrc@act.buaa.edu.cn},
    Yongyi Mao\textsuperscript{2},
    Jie Xu\textsuperscript{3}
}
\affiliations{
    \textsuperscript{\rm 1}SKLSDE, School of Computer Science and Engineering, Beihang University, Beijing, China \\
         \textsuperscript{\rm 2}School of Electrical Engineering and Computer Science, University of Ottawa, Ottawa, Canada \\
         \textsuperscript{\rm 3}Department of Computer Science, University of Leeds, UK \\
    chenjf@act.buaa.edu.cn, zhangrc@act.buaa.edu.cn, ymao@uottawa.ca, j.xu@leeds.ac.uk
}

\usepackage{bibentry}

\begin{document}

\maketitle


\begin{abstract}
Few-shot text classification has recently been promoted by the meta-learning paradigm which aims to identify target classes with knowledge transferred from source classes with sets of small tasks named episodes. Despite their success, existing works building their meta-learner based on Prototypical Networks are unsatisfactory in learning discriminative text representations between similar classes, which may lead to contradictions during label prediction. In addition, the task-level and instance-level overfitting problems in few-shot text classification caused by a few training examples are not sufficiently tackled. In this work, we propose a contrastive learning framework named ContrastNet to tackle both discriminative representation and overfitting problems in few-shot text classification. ContrastNet learns to pull closer text representations belonging to the same class and push away text representations belonging to different classes, while simultaneously introducing unsupervised contrastive regularization at both task-level and instance-level to prevent overfitting. Experiments on 8 few-shot text classification datasets show that ContrastNet outperforms the current state-of-the-art models.
\end{abstract}
\section{Introduction}
Building a human-like learning system that has the ability to quickly learn new concepts from scarce experience is one of the targets in modern Artificial Intelligence (AI) communities. Meta-learning or referred to as learning to learn is such a few-shot learning paradigm that aims to mimics human abilities to learn from different small tasks (or episodes) of source classes in the training set and generalize to unseen tasks of target classes in the test set. Meta-learning has been extensively studied in image classification and achieve remarkable successes~\cite{Vinyals:16, Snell:17, Finn:17, Sung:18, Hou:19, Tseng:20, Liu:21, Gao:21}. The effectiveness in image classification motivates the recent application of meta-learning to few-shot text classification~\cite{Yu:18, Geng:19, Geng:20, Bao:20, Han:21}.

\begin{figure}[ht]
	\begin{center}
		\begin{tabular}{cc}
		\hspace{-.4cm}
			\scalebox{0.47}{
		    \includegraphics{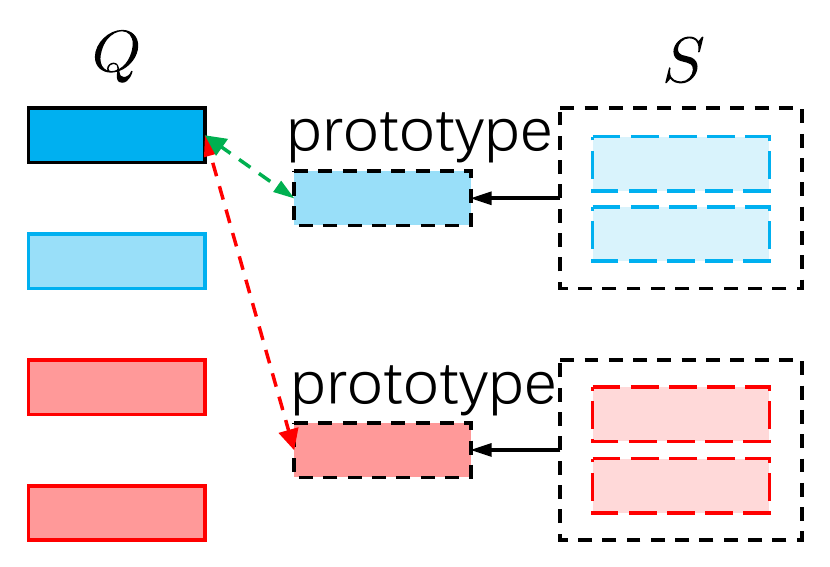}
			}&
		\hspace{-.25cm}
			\scalebox{0.47}{
			\includegraphics{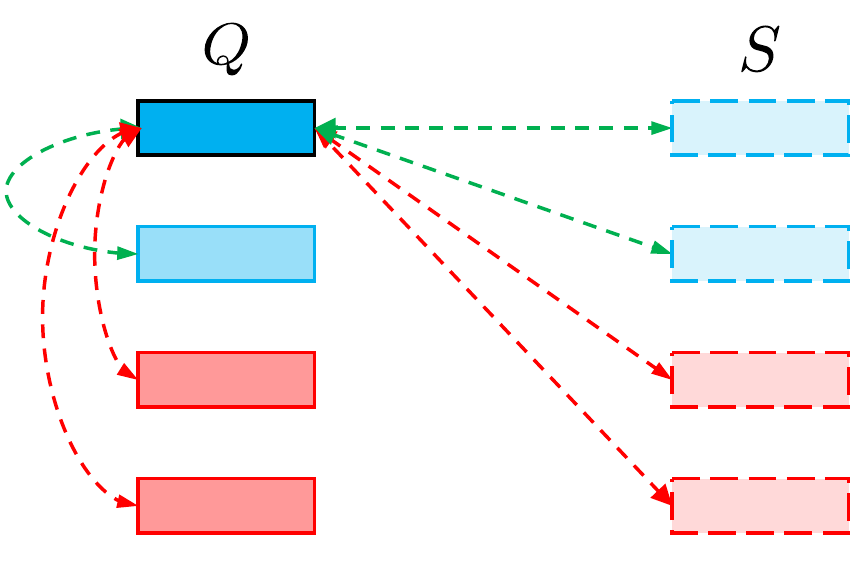}
			}\\
		\hspace{-.3cm}
		(a) Prototypical Networks & 
		\hspace{0.2cm}
		(b) ContrastNet
		\end{tabular}
	\end{center}
	\caption{The learning strategies of Prototypical Network and proposed ContrastNet. Q and S respectively denote the query set and support set. The rectangles with different colors denote text representations from different classes. The green and red dashed arrow lines respectively indicate pulling closer and pushing away the representations. Picture (a) shows that Prototypical Networks learn to align a given query-text representations to prototypes computed by support-text representations. Picture (b) shows that ContrastNet learns to pull closer the given query-text representation with text representations belonging to the same class and push away text representations with different classes.}
	\vspace{-.4cm}
	\label{fig:intro}
\end{figure}

One of the metric-based meta-learning methods that has been widely studied and shown effectiveness in few-shot learning is Prototypical Networks~\cite{Snell:17}. As shown in Figure~\ref{fig:intro} (a), at each episode, Prototypical Networks first compute the prototype for each class using the text representations in the support set, then align each text representation in the query set to the prototypes under some measurement, e.g., Euclidean distance. This learning strategy allows the meta-learner to perform few-shot text classification by simply learning the representations of the texts.  However, as the model design in Prototypical Networks ignores the relationships among the texts in the query set, the  discrimination among the query-text representations is not guaranteed, which may lead to difficulty in prediction when two text representations in the query set are very similar but they belong to different classes. Such similar texts with different classes are common because real-world few-shot text classification tasks may involve fine-grained classes with very similar semantics. For example, in intent classification, the sentences ``who covered the song one more cup of coffee" with intent {\em music-query} and ``play the song one more cup of coffee" with intent {\em music-play} may produce similar text representations but they belong to different intents. When these two sentences are sampled in the same query set, they are hard to distinguish from each other and bring about contradiction in prediction because they will obtain similar measurements aligning to each prototype, thus may lead to misclassification. 

To tackle the above issue caused by similar text representations of similar classes, we propose a few-shot text classification framework ContrastNet that encourages learning discriminative text representations via contrastive learning, motivated by its successful application in few-shot image classification~\cite{Gao:21, LuoX:21, Chen:21, Majumder:21, Liu:21}. As shown in Figure~\ref{fig:intro} (b), in ContrastNet, the text representations are learned free from the prototypes by pulling closer a text representation with text representations belonging to the same class and push away text representations with different classes from both query and support set. In this way, when two texts with similar semantics from different classes are sampled in the same query set, they are forced to produce discriminative representations by the contrastive loss, thus alleviate the contradictions during prediction.

Another challenge in few-shot text classification is that the models are prone to overfit the source classes based on the biased distribution formed by a few training examples ~\cite{Yang:21, Dopierre:21}. The authors of \cite{Yang:21} propose to tackle the overfitting problem in few-shot image classification by training with additional instances generated from calibrated distributions. In few-shot text classification, PROTAUGMENT~\cite{Dopierre:21} introduce an unsupervised cross-entropy loss with unlabeled instances to prevent the model from overfitting the source classes. Although successful, these approaches only tackle the instance-level overfitting. In this paper, we argue that the overfitting may also occur at task-level because not only the text instances from target classes but also the way they are combined as tasks are unavailable during training.

We incorporate two unsupervised contrastive losses as the regularizers upon the basic supervised contrastive learning model to alleviate the instance-level and task-level overfitting problems. Specifically, the representations of randomly sampled tasks from source classes and the representations of randomly sampled unlabeled texts with their augmentations are taken to form a task-level contrastive loss and an instance-level contrastive loss in an unsupervised manner, respectively. The unsupervised task-level and instance-level contrastive losses force the representations of different tasks and different unlabeled texts to be separated from each other in their representation space. We hope this separation to pull the task and instance representations of target classes away from the task and instance representations of source classes, thus alleviate the overfitting problems. 

To summarize, our work makes the following contributions. (1) We propose a few-shot text  classification framework ContrastNet that learns discriminative text representations via contrastive learning to reduce contradictions during prediction caused by similar text representations of similar classes. (2) We introduce two unsupervised contrastive losses as regularizers upon the basic supervised contrastive representation model, which alleviate the task-level and instance-level overfitting in few-shot text classification by learning separable task representations and instance representations. (3) We conduct experiments on 8 text classification datasets and show that ContrastNet outperforms the start-of-the-arts. Additional analysis on the results comparing to Prototypical Networks shows that ContrastNet effectively learns discriminative text representations and alleviates the task-level and instance-level overfitting problems. 

\section{Problem Formulation}
The meta-learning paradigm of few-shot text classification aims to transfer knowledge learned from sets of small tasks (or episodes) of source classes to target classes which are unseen during training.

Formally, let $\mathcal{Y}_{train}$, $\mathcal{Y}_{val}$ and $\mathcal{Y}_{test}$ denote the disjoint set of training classes, validation classes and test classes, i.e., they have no overlapping classes. At each episode, a task composed of a support set $\mathcal{S}$ and a query set $\mathcal{Q}$ is drawn from the dataset of either $\mathcal{Y}_{train}$, $\mathcal{Y}_{val}$ and $\mathcal{Y}_{test}$ during training, validation or test. In an episode of a $n$-way $k$-shot text classification problem, $n$ classes are sampled from corresponding class set; for each of the $n$ classes, $k$ labeled texts are sampled to compose the support set, and $m$ unlabeled texts are sampled to compose the query set. 

For convenience, we use a pair $(x^{s}_{i}, y^{s}_{i})$ to denote the $i^{\rm th}$ item of total $n \times k$ items in the support set $\mathcal{S}$ and $x^{q}_{j}$ denotes the $j^{\rm th}$ text instance of total $n \times m$ instances in the query set $\mathcal{Q}$. For the text instance $x^{q}_{j}$, we denote its class label as $y^{q}_{j}$. A meta-learner is trained on such small tasks that attempts to classify the texts in the query set $\mathcal{Q}$ on the basis of few labeled texts in the support set $\mathcal{S}$.

\section{Methodology}
\begin{figure*}[ht]
    \centering
	\scalebox{0.5}{
	    \includegraphics{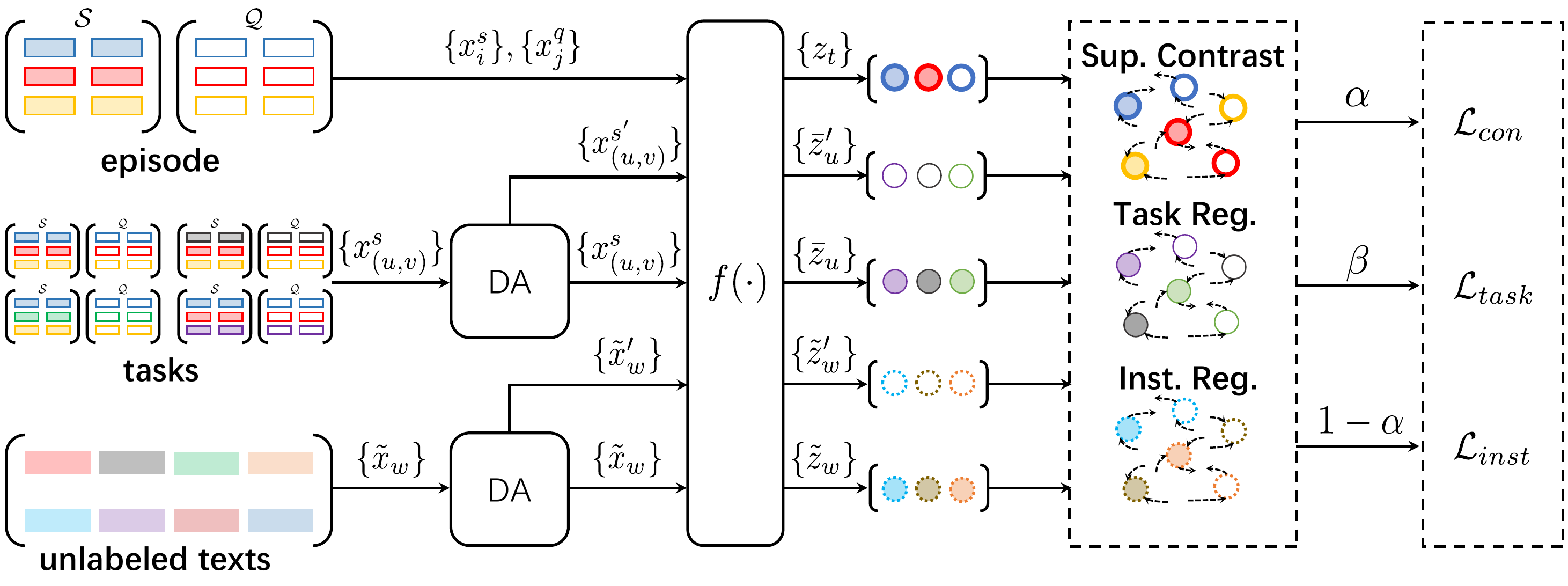}
	}
	\caption{The overall model structure of ContrastNet. The DA blocks represent data augmentation.}
	\label{fig:structure}
\end{figure*}

Our ContrastNet combines BERT text encoder and supervised contrastive learning to learn discriminative text representations and incorporates the task-level and instance-level unsupervised contrastive regularization to alleviate the overfitting problems. The overall model structure of ContrastNet is shown in Figure~\ref{fig:structure}. All notations in Figure~\ref{fig:structure} will be defined in the rest of this section.


\subsection{Supervised Contrastive Text Representation}

\subsubsection{Text Encoder}
In metric-based few-shot text classification, a text encoder is needed to map the raw text onto a vector space where the metrics (or measurements) between texts can be computed. The pre-trained language models, such as BERT, have recently been employed as text encoders to obtain text representations and achieve promising results. Following previous works in few-shot text classification~\cite{Bansal:20, Luo:21, Dopierre:21}, we also utilize BERT to represent the texts. Specifically, BERT takes a text $x$ composed of a list of tokens as input, and output a hidden-state vector for each of the tokens; we take the hidden-state vector corresponding to the {\em CLS} token as the text representation of $x$. For later use, we denote the BERT text representation module as $f(\cdot)$ and denote all of its parameters as $\theta$.

\subsubsection{Supervised Contrastive Learning}
Our few-shot learning framework is also a metric-based approach, but different from Prototypical Networks that align query texts with prototypes, we optimize the measurement free of prototypes, by learning to align two text representations using supervised contrastive learning. It pulls closer the text representations belonging to the same class and pushes away text representations belonging to different classes among texts from both query and support sets. 

The model design of our supervised contrastive learning is based on the ``batch contrastive learning" framework~\cite{Chen:20} and the supervised contrastive learning strategy in \cite{Khosla:20}. Specifically, given the support set $\mathcal{S}$ and query set $\mathcal{Q}$ in an episode, we combine the $n \times k$ text instances $\{x^s_{i}\}$ in $\mathcal{S}$ and the $n \times m$ text instances $\{x^q_{j}\}$ in $\mathcal{Q}$ as a training batch $\mathcal{B}=\{ x_1, x_2, \cdots, x_{n(k+m)} \}$, where 

\begin{equation}
\begin{split}
x_{t}=\left\{\begin{matrix}
x^s_{t}, & t \leqslant nk\\ 
x^q_{t-nk}, & t > nk
\end{matrix}\right. 
\end{split}
\label{eq:batch}
\end{equation}

For each $x_t \in \mathcal{B}$, we denote its label as $y_t$ and denote its representation transformed by $f(\cdot)$ as $z_t$. The matched text-instance pairs and unmatched text-instance pairs in the batch is identified based on their labels. Let $c=k+m-1$ be the number of text instances in $\mathcal{B}$ which has the same label as $x_t$. The text representations can then be optimized by following supervised contrastive loss
\begin{equation}
\begin{split}
\mathcal{L}_{con} \!=\! -\!\!\sum\limits_{\mathbf{x}_t\in \mathcal{B}} \!\frac{1}{c} \log\frac{\sum\limits_{y_r=y_t}\!\!\!\exp(z_t\!\cdot\! z_r/\tau)}{\!\!\!\sum\limits_{y_{r}=y_t}\!\!\!\exp(z_t\!\cdot\! z_r/\tau)\!+\!\!\!\!\!\sum\limits_{y_{r'} \neq y_t}\!\!\!\exp(z_t\!\cdot\! z_{r'}/\tau)}
\end{split}
\label{eq:super}
\end{equation}
where the inner product is used as the similarity measurement of two text representations, and $\tau$ is a temperature factor that scales the inner products. 

The supervised contrastive loss in Equation (\ref{eq:super}) encourages each representation $z^q$ of query-text $x^q \in \mathcal{Q}$ to locate near the query-text representations that have the same class label with $x^q$ and distant from the query-text representations that have different class labels with $x^q$, thus increase the discrimination of query-text representations between different classes and alleviate the contradictions in label prediction. 

\subsection{Unsupervised Contrastive Regularization}
To tackle the overfitting problems caused by a few training examples in few-shot text classification, we propose to train the supervised contrastive representation model under the regularization of a task-level unsupervised contrastive loss and an instance-level unsupervised contrastive loss. 

\subsubsection{Data Augmentation}
Data augmentation has shown to be essential in boosting contrastive learning~\cite{Chen:20, Tian:20, Kalantidis:20, You:20, Cai:20, GaoT:21}. However, data augmentation of text is still an open challenge. Among the direction of textual data augmentation, the EDA~\cite{Wei:19} may alter the text purport~\cite{Sun:21} and the back translation fails to provide diverse augmentations~\cite{Dopierre:21}. The recent work PROTAUGMENT~\cite{Dopierre:21} propose a short-text paraphrasing model that produces diverse paraphrases of the original text as data augmentations. As the data augmentations of PROTAUGMENT have shown to be effective in few-shot text classification, we apply PROTAUGMENT to generate data augmentations of the texts in our unsupervised contrastive learning.

\subsubsection{Task-level Contrastive Regularization}
In few-shot text classification, the seen tasks are sampled from the source classes $\mathcal{Y}_{train}$, while the unseen tasks sampled from the target classes $\mathcal{Y}_{test}$ are unavailable during training. Therefore, the models tend to overfit the seen tasks if trained without constraint and degrade performance when it generalizes to unseen tasks. Our solution to this problem is to constrain the model with an unsupervised contrastive loss built upon randomly sampled tasks and their data augmentations.

Specifically, at each episode, we randomly sample $N_{task}$ tasks $\{ (\mathcal{Q}_1, \mathcal{S}_1), (\mathcal{Q}_2, \mathcal{S}_2), \cdots, (\mathcal{Q}_{N_{task}}, \mathcal{S}_{N_{task}}) \}$ from the source classes $\mathcal{Y}_{train}$, and we use $x^s_{(u, v)}$, $x^{s'}_{(u, v)}$ and $z^s_{(u, v)}$ to respectively denote the $v^{\rm th}$ text instance, its text augmentation and its text representation in support set $\mathcal{S}_u$ of the $u^{\rm th}$ task. The representation $\bar{z}_u$ of the $u^{\rm th}$ task can simply be calculated as the mean embedding of all text instances in $\mathcal{S}_u$. To obtain the data augmentation of the $u^{\rm th}$ task, we replace the text instances in $\mathcal{S}_u$ with their corresponding text augmentations, and similarly, we compute the mean embedding $\bar{z}'_u$ of these text augmentations as the data augmentation of the $u^{\rm th}$ task. We combine all $\bar{z}_u$ and $\bar{z}'_u$ as a training batch $\{\bar{z}_u\}$ of $2N_{task}$ elements and use $\bar{z}'_u$ denotes the matched element of $\bar{z}_u$ in $\{\bar{z}_u\}$. The task-level contrastive regularization loss is
\begin{equation}
\begin{split}
\mathcal{L}_{task} \!=\! -\!\!\!\!\sum_{u=1}^{2N_{task}}\!\!\!\log\frac{\exp(\bar{z}_u\!\cdot\! \bar{z}'_u/\tau)}{\exp(\bar{z}_u\!\cdot\! \bar{z}'_u/\tau)+\!\!\!\!\!\sum\limits_{\bar{z}_{u'} \neq \bar{z}'_u}\!\!\!\exp(\bar{z}_u\!\cdot\! \bar{z}_{u'}/\tau)}
\end{split}
\label{eq:task}
\end{equation}
The unsupervised contrastive loss in Equation (\ref{eq:task}) forces the representations of different tasks (or compositions of classes) to be separated from each other. Separation of tasks encourages the separation of classes between tasks.  This separation urges the representations of the unseen tasks to locate distant from the seen tasks, thus alleviate the task-level overfitting problem.

\subsubsection{Instance-level Contrastive Regularization}
The instance-level overfitting in few-shot text classification is not entirely unknown to the research community. The PROTAUGMENT introduces an unsupervised cross-entropy loss upon Prototypical Networks, which encourages the representation of each unlabeled text being closer to its augmentations’ prototype and distant from the prototypes of other unlabeled texts. In this work, we build a different instance-level unsupervised loss that serves as a regularizer of the supervised contrastive text representation model. Our objective is to prevent instance-level overfitting by learning separable text representations between source and target classes. To that end, we introduce the instance-level unsupervised contrastive regularization.

Specifically, at each training episode, we randomly sample $N_{inst}$ unlabeled text instances $\{ \tilde{x}_{1}, \tilde{x}_{2}, \cdots, \tilde{x}_{N_{inst}} \}$. Let $\tilde{x}'_{w}$ denote the data augmentation of text instance $\tilde{x}_{w}$; $\tilde{z}_{w}$ and $\tilde{z}'_{w}$ denote the text representation of $\tilde{x}_{w}$ and $\tilde{x}'_{w}$, respectively. We combine all $\tilde{x}_{w}$ and $\tilde{x}'_{w}$ as a training batch $\{\tilde{x}_{w}\}$ of $2N_{inst}$ elements and use $\tilde{x}'_{w}$ to denote the matched element of $\tilde{x}_{w}$ in $\{\tilde{x}_{w}\}$. The instance-level contrastive regularization loss is
\begin{equation}
\begin{split}
\mathcal{L}_{inst} \!=\! -\!\!\!\!\sum_{w=1}^{2N_{inst}}\!\!\!\log\frac{\exp(\tilde{z}_w\!\!\cdot\! \tilde{z}'_w/\tau)}{\exp(\tilde{z}_w\!\!\cdot\! \tilde{z}'_w/\tau)+\!\!\!\!\!\!\sum\limits_{\tilde{z}_{w'} \neq \tilde{z}'_w}\!\!\!\!\exp(\tilde{z}_w\!\!\cdot\! \tilde{z}_{w'}/\tau)}
\end{split}
\label{eq:inst}
\end{equation}
The unsupervised contrastive loss in Equation (\ref{eq:inst}) encourages different text representations locating distant from each other, which prevents the text representations of target classes from being too closer to text representations of source classes, thus alleviate the instance-level overfitting.

\subsection{Objective and Prediction}

\subsubsection{Overall Objective}
During training, we combine the loss $\mathcal{L}_{con}$ of the supervised contrastive text representation learning model with the unsupervised regularization losses $\mathcal{L}_{inst}$ at the instance-level and $\mathcal{L}_{task}$ at the task-level. The overall objective is
\begin{equation}
\begin{split}
\mathcal{L} = \alpha \mathcal{L}_{con} + (1-\alpha) \mathcal{L}_{inst} + \beta \mathcal{L}_{task}
\end{split}
\label{eq:loss}
\end{equation}
where $\alpha$ and $\beta$ are hyper-parameters that indicate the weights on the loss of supervised contrastive learning and task-level unsupervised regularization loss, respectively. The overall model can be optimized using stochastic gradient descent (SGD) methods.

\subsubsection{Label prediction} 
As the text representations in ContrastNet are learned free of prototypes, the label prediction setup in Prototypical Networks that align the query text to prototypes with the maximum measurement is no longer appropriate to ContrastNet. A natural label prediction setup for ContrastNet is to infer the label of a query text by comparing its representation with text representations from the support set. In this work, we adopt the Nearest Neighbor classifier as such a label prediction setup. Specifically, given a query text $x^q \in \mathcal{Q}$, we first obtain its representation $f(x^q)$ and representations of all texts in the support set $\{ f(x^s_{i}) \}$, then the label of query text $x^q$ is determined as the label $y^s_{i}$ of the support-text whose representation $f(x^s_{i})$ has the maximum inner product with $f(x^q)$. Let $y^s_{i^*}$ be the predicted label, then the process to find $i^*$ can be formulated as
\begin{equation}
\begin{split}
i^* = \arg\max_{i}f(x^q)\cdot f(x^s_{i})
\end{split}
\label{eq:predict}
\end{equation}

\section{Experiments}

\subsection{Datasets}
We evaluate our few-shot text classification models on 8 text classification datasets, including $4$ intent classification datasets: {\bf Banking77}~\cite{Casanueva:20}, {\bf HWU64}~\cite{Liu:19}, {\bf Clinic150}~\cite{Larson:19}, {\bf Liu57}~\cite{LiuX:19} and $4$ news or review classification datasets: {\bf HuffPost}~\cite{Bao:20}, {\bf Amazon}~\cite{He:16}, {\bf Reuters}~\cite{Bao:20}, {\bf 20News}~\cite{Lang:95}. The statistics of the datasets are shown in Table~\ref{tab:statics}.

\begin{table*}[ht]
	\centering
    \scalebox{1.0}{
	\begin{tabular}{ccccc}
		\toprule
		dataset & train/valid/test classes & sentences & avg\_sent\_class & avg\_tok\_sent\\
		\midrule
		Banking77 & 25/25/27 & $13,083$ & $170$ & $12$ \\
		HWU64 & 23/16/25 & $11,036$ & $172$ & $7$ \\
		Clinic150 & 50/50/50 & $22,500$ & $150$ & $9$ \\
		Liu & 18/18/18 & $25,478$ & $472$ & $8$ \\
		HuffPost & 20/5/16 & $36,900$ & $900$ & $11$ \\
		Amazon & 10/5/9 & $24,000$ & $1000$ & $140$ \\
		Reuters & 15/5/11 & $620$ & $20$ & $168$ \\
		20News & 8/5/7 & $18,820$ & $941$ & $340$ \\
		\bottomrule
	\end{tabular}}
	\caption{The statistics of few-shot text classification datasets.}
	\label{tab:statics}
\end{table*}

\subsubsection{Intent Classification Datasets} 
The Banking77 dataset is a fine-grained intent classification dataset specific to a single banking domain, which includes $13,083$
user utterances divided into $77$ different
intents. The HWU64 dataset is also a fine-grained intent classification dataset but the classes are across multi-domain, which contains $11,036$ user utterances with $64$ user intents from $21$ different domains. The Clinic150 intent classification dataset contains $22,500$ user utterances equally distributed in $150$ intents. Following \cite{Mehri:20, Dopierre:21}, we only keep the 150 intent labels and discard the out-of-scope intent labels in our experiment. Liu57 is a highly imbalanced intent classification dataset collected on Amazon Mechanical Turk, which is composed of $25,478$ user utterances from 54 classes. 

\subsubsection{News or Review Classification Datasets} 
The HuffPost dataset is a news classification dataset with $36,900$ HuffPost news headlines with $41$ classes collected from the year 2012 to 2018. The Amazon dataset is a product review classification dataset including $142.8$
million reviews with $24$ product categories from the year 1996 to 2014. We use the subset provided by \cite{Han:21}, in which each class contains $1000$ sentences. The Reuters dataset is collected from Reuters newswire in 1987. Following \cite{Bao:20}, we only use
$31$ classes and remove the multi-labeled articles. The 20News dataset is a news classification dataset, which contains $18,820$ news documents from $20$ news groups. 

\subsection{Experimental Settings}
We evaluate our models on typical $5$-way $1$-shot and $5$-way $5$-shot settings. Following the setup in \cite{Dopierre:21}, we report the average accuracy over $600$ episodes sampled from the test set for intent classification datasets; and following \cite{Han:21}, we report the average accuracy over $1000$ episodes sampled from the test set for news or review classification datasets. We run each experimental setting 5 times. For each run, the training, validation, and testing classes are randomly re-split.   

We implement the proposed models using Pytorch deep learning framework \footnote{Our code and data are available at: https://github.com/BDBC-KG-NLP/AAAI2022\_ContrastNet.}. On the 4 intent classification datasets, we use their respective pre-trained BERT-based language model provided in \cite{Dopierre:21} as the encoders for text representation. For the news or review classification datasets, we use the pure pre-trained {\tt bert-base-uncased} model as the encoder for text representation. We use EDA to augment texts in Amazon, Reuters and 20News because they are long sequences unsuitable for PROTAUGMENT. For each episode during training, we randomly sample $10$ tasks and $10$ unlabeled texts to calculate the task-level contrastive regularization loss and instance-level contrastive regularization loss. The temperature factors of loss $\mathcal{L}_{con}$, $\mathcal{L}_{task}$ and $\mathcal{L}_{inst}$ are set to $5.0$, $7.0$ and $7.0$, respectively. The loss weight $\alpha$ is initialized to $0.95$ and decrease during training using the loss annealing strategy~\cite{Dopierre:21}, and the loss weight $\beta$ is set to $0.1$. We optimize the models using Adam~\cite{Kingma:15} with an initialized learning rate of $1\times e^{-6}$. All the hyper-parameters are selected by greedy search on the validation set. All experiments are run on a single NVIDIA Tesla V100 PCIe 32GB GPU.

\subsection{Baseline Models}
We compare the proposed few-shot text classification models with following baselines:

\noindent {\bf Prototypical Networks} 
This model is a metric-based meta-learning
method for few-shot classification proposed in \cite{Snell:17}, which learns to align query instances with class prototypes.

\noindent {\bf MAML} 
This model is proposed in \cite{Finn:17}, which learns to rapidly adapt to new tasks by only few gradient steps.

\noindent {\bf Induction Networks}
This model is proposed in \cite{Geng:19}, which introduces dynamic routing algorithm to learn the class-level representation.

\noindent {\bf HATT} 
This model is proposed in \cite{Gao:19}, which extends the prototypical networks by incorporating a hybrid attention mechanism.

\noindent {\bf DS-FSL} 
This model is proposed in \cite{Bao:20}, which aims to extract more transferable features by mapping the distribution signatures to attention scores.

\noindent {\bf MLADA} 
This model is proposed in \cite{Han:21}, which adopts adversarial networks to improve the domain adaptation ability of meta-learning.

\noindent {\bf PROTAUGMENT} 
This model is proposed in \cite{Dopierre:21}, which utilizes a short-texts paraphrasing model to generate data augmentation of texts and builds an instance-level unsupervised loss upon the prototypical networks. We also report its two improved versions with different word masking strategies, i.e., PROTAUGMENT (unigram) and PROTAUGMENT (bigram).

\begin{table*}[ht]
	\centering
    \scalebox{0.9}{
	\begin{tabular}{ccccccccccc}
		\toprule
		\multirow{2}*{\textbf{Method}}&\multicolumn{2}{c}{\textbf{Banking77}}&\multicolumn{2}{c}{\textbf{HWU64}}&\multicolumn{2}{c}{\textbf{Liu}}&\multicolumn{2}{c}{\textbf{Clinic150}}&\multicolumn{2}{c}{\textbf{Average}}\\
		\cmidrule(lr){2-3} \cmidrule(lr){4-5} \cmidrule(lr){6-7} \cmidrule(lr){8-9} \cmidrule(lr){10-11}&
		1-shot & 5-shot & 1-shot & 5-shot & 1-shot & 5-shot & 1-shot & 5-shot & 1-shot & 5-shot \\
		\midrule
		Prototypical Networks & 86.28 & 93.94 & 77.09 & 89.02 & 82.76 & 91.37 & 96.05 & 98.61 & 85.55$\pm$2.20 & 93.24$\pm$1.22 \\
		PROTAUGMENT & 86.94 & 94.50 & 82.35 & 91.68 & 84.42 & 92.62 & 94.85 & 98.41 & 87.14$\pm$1.36 & 94.30$\pm$0.60 \\
		PROTAUGMENT (bigram) & 88.14 & 94.70 & 84.05 & 92.14 & 85.29 & 93.23 & 95.77 & 98.50 & 88.31$\pm$1.43 & 94.64$\pm$0.59 \\
		PROTAUGMENT (unigram) & 89.56 & 94.71 & 84.34 & 92.55 & 86.11 & 93.70 & 96.49 & {\bf 98.74} & 89.13$\pm$1.13 & 94.92$\pm$0.57 \\
		\midrule
		{\bf ContrastNet} ($\mathcal{L}_{task} \& \mathcal{L}_{inst}$ /o) & 88.53 & 95.22 & 84.62 & 91.93 & 80.53 & 93.47 & 94.29 & 98.09 & 86.99$\pm$1.57 & 94.68$\pm$0.74 \\
		{\bf ContrastNet} ($\mathcal{L}_{inst}$ /o) & 89.75 & 95.36 & 85.14 & 91.69 & {\bf 86.79} & 93.28 & 96.32 & 98.25 & 89.50$\pm$1.30 & 94.65$\pm$0.64 \\
		{\bf ContrastNet} & {\bf 91.18} & {\bf 96.40} & {\bf 86.56} & {\bf 92.57} & 85.89 & {\bf 93.72} & {\bf 96.59} & 98.46 & {\bf 90.06}$\pm${\bf 1.02} & {\bf 95.29}$\pm${\bf 0.53} \\
		\bottomrule
	\end{tabular}
	}
	\caption{The 5-way 1-shot and 5-way 5-shot text classification results on the Banking77, HWU64, Liu and Clinic150 intent classification datasets. The ContrastNet ($\mathcal{L}_{task} \& \mathcal{L}_{inst}$ /o) model denote the ContrastNet only using supervised contrastive text representation without any unsupervised regularization and the ContrastNet ($\mathcal{L}_{inst}$ /o) model denotes the ContrastNet with only task-level unsupervised regularization. We compute the mean and the standard deviation over 5 runs with different class splitting. The {\bf Average} denotes the averaged mean and standard deviation over all datasets for each setting of each model.}
	\label{tab:resIntent}
\end{table*}

\begin{table*}[ht]
	\centering
    \scalebox{0.9}{
	\begin{tabular}{ccccccccccc}
		\toprule
		\multirow{2}*{\textbf{Method}}&\multicolumn{2}{c}{\textbf{HuffPost}}&\multicolumn{2}{c}{\textbf{Amazon}}&\multicolumn{2}{c}{\textbf{Reuters}}&\multicolumn{2}{c}{\textbf{20News}}&\multicolumn{2}{c}{\textbf{Average}}\\
		\cmidrule(lr){2-3} \cmidrule(lr){4-5} \cmidrule(lr){6-7} \cmidrule(lr){8-9} \cmidrule(lr){10-11}&
		1-shot & 5-shot & 1-shot & 5-shot & 1-shot & 5-shot & 1-shot & 5-shot & 1-shot & 5-shot \\
		\midrule
		MAML & 35.9 & 49.3 & 39.6 & 47.1 & 54.6 & 62.9 & 33.8 & 43.7 & 40.9 & 50.8 \\
		Prototypical Networks & 35.7 & 41.3 & 37.6 & 52.1 & 59.6 & 66.9 & 37.8 & 45.3 & 42.7 & 51.4 \\
		Induction Networks & 38.7 & 49.1 & 34.9 & 41.3 & 59.4 & 67.9 & 28.7 & 33.3 & 40.4 & 47.9 \\
		HATT & 41.1 & 56.3 & 49.1 & 66.0 & 43.2 & 56.2 & 44.2 & 55.0 & 44.4 & 58.4 \\
		DS-FSL & 43.0 & 63.5 & 62.6 & 81.1 & 81.8 & 96.0 & 52.1 & 68.3 & 59.9 & 77.2 \\
		MLADA & 45.0 & 64.9 & 68.4 & {\bf 86.0} & 82.3 & {\bf 96.7} & 59.6 & 77.8 & 63.9 & 81.4 \\
		\midrule
		{\bf ContrastNet} ($\mathcal{L}_{task} \& \mathcal{L}_{inst}$ /o) & 52.74 & 63.59 & 74.70 & 84.47 & 83.74 & 93.28 & 70.61 & 80.04 & 70.45$\pm$3.28 & 80.35$\pm$3.32 \\
		{\bf ContrastNet} ($\mathcal{L}_{inst}$ /o) & 52.85 & 64.88 & 75.33 & 84.21 & 85.10 & 93.65 & 70.35 & 80.19 & 70.91$\pm$3.00 & 80.73$\pm$2.79 \\
		{\bf ContrastNet} & {\bf 53.06} & {\bf 65.32} & {\bf 76.13} & 85.17 & {\bf 86.42} & 95.33 & {\bf 71.74} & {\bf 81.57} & {\bf 71.84}$\pm${\bf 2.81} & {\bf 81.85}$\pm${\bf 2.03} \\
		\bottomrule
	\end{tabular}
	}
	\caption{The 5-way 1-shot and 5-way 5-shot text classification results on the HuffPost, Amazon, Reuters and 20News datasets. }
	\label{tab:resnews}
\end{table*}

\subsection{Few-shot Text Classification Results}

\subsubsection{Main Results} 
The few-shot text classification results in 5-way 1-shot and 5-way 5-shot settings are shown in Table~\ref{tab:resIntent} and Table~\ref{tab:resnews}. We take the results of baseline models from \cite{Dopierre:21} for the 4 intent classification datasets and from \cite{Han:21} for the 4 news and review classification datasets.
The current state-of-the-art (SOTA) models on the 4 intent classification datasets and the 4 news and review classification datasets are PROTAUGMENT (unigram) and MLADA, respectively. From Table~\ref{tab:resIntent} and Table~\ref{tab:resnews}, we observe that ContrastNet achieves the best average results in both 5-way 1-shot setting and 5-way 5-shot setting on all datasets. ContrastNet builds itself as the new SOTA in both 5-way 1-shot and 5-way 5-shot settings on all datasets, except in 5-way 1-shot setting of Liu and 5-way 5-shot setting of Clinic150, Amazon, Reuters. ContrastNet also achieves significantly higher accuracy than the current SOTA models on most of the few-shot text classification datasets in 5-way 1-shot setting.
These significant improvements suggest that learning discriminative text representations using the supervised contrastive learning with task-level and instance-level regularization can efficiently raise the few-shot text classification performance. 

\subsubsection{Ablation Study} 
We consider two ablated models of ContrastNet: ContrastNet ($\mathcal{L}_{inst}$ /o) that removes the instance-level regularization loss from ContrastNet and ContrastNet ($\mathcal{L}_{task} \& \mathcal{L}_{inst}$ /o) that removes both instance-level and task-level regularization losses from ContrastNet. From the ablation results in Table \ref{tab:resIntent} and Table~\ref{tab:resnews}, we observe that ContrastNet ($\mathcal{L}_{inst}$ /o) improves few-shot text classification performance upon ContrastNet ($\mathcal{L}_{task} \& \mathcal{L}_{inst}$ /o); ContrastNet further promotes ContrastNet ($\mathcal{L}_{inst}$ /o). These results demonstrate the effectiveness of task-level and instance-level regularization in promoting the basic supervised contrastive representation model. The ContrastNet ($\mathcal{L}_{task} \& \mathcal{L}_{inst}$ /o) with the pure supervised contrastive loss already outperforms Prototypical Networks on all datasets except Liu and Clinic150, which suggests the power of supervised contrastive learning in producing discriminative text representations tand improving the accuracy.

\subsection{Results Analysis Based on Similar Classes}
\subsubsection{Visualizing Text Representations of Similar Classes} 
\begin{figure}[ht]
	\begin{center}
		\begin{tabular}{cc}
		\hspace{-.25cm}
			\scalebox{0.29}{
		    \includegraphics{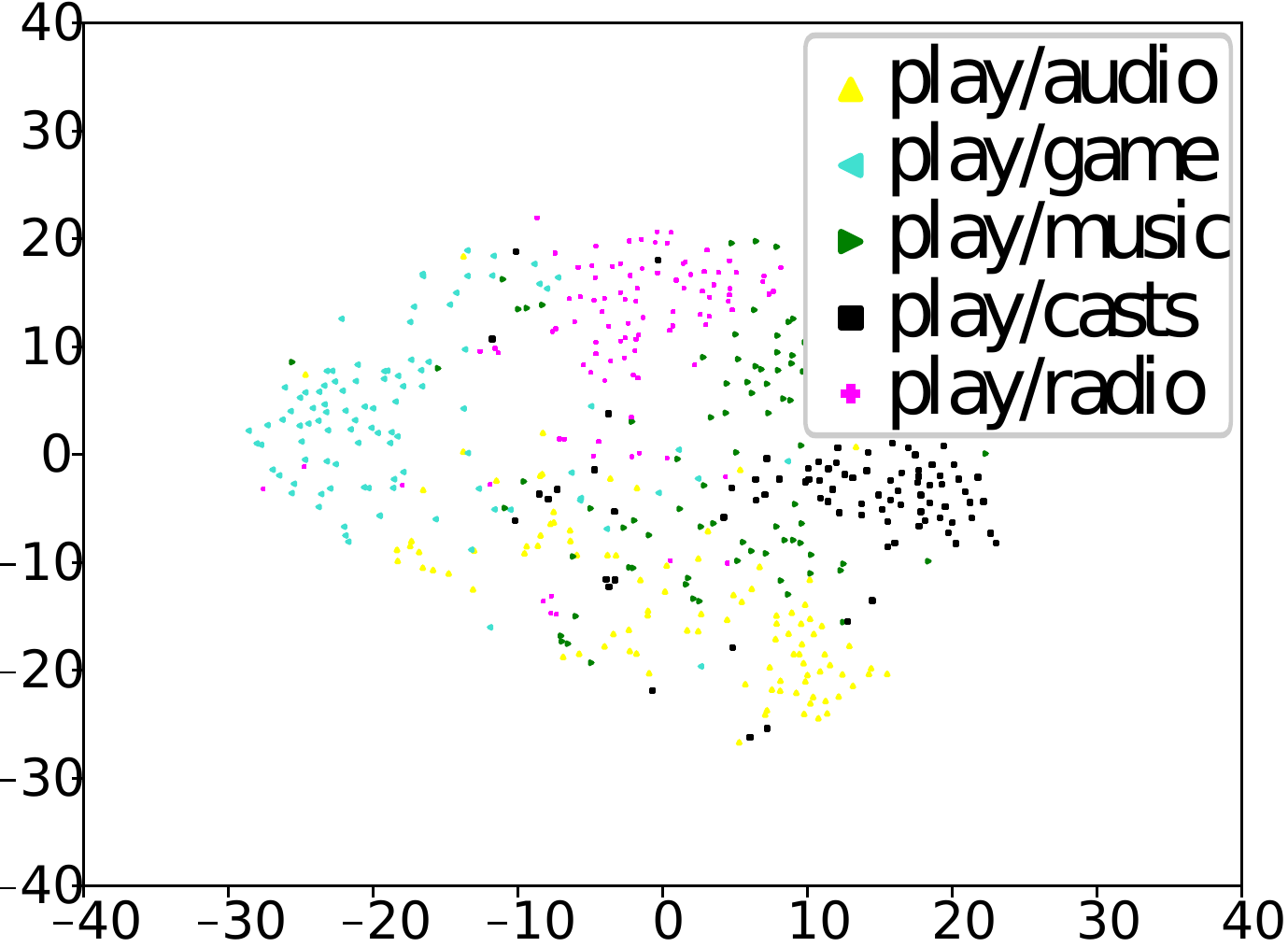}
			}&
		\hspace{-.55cm}
			\scalebox{0.29}{
			\includegraphics{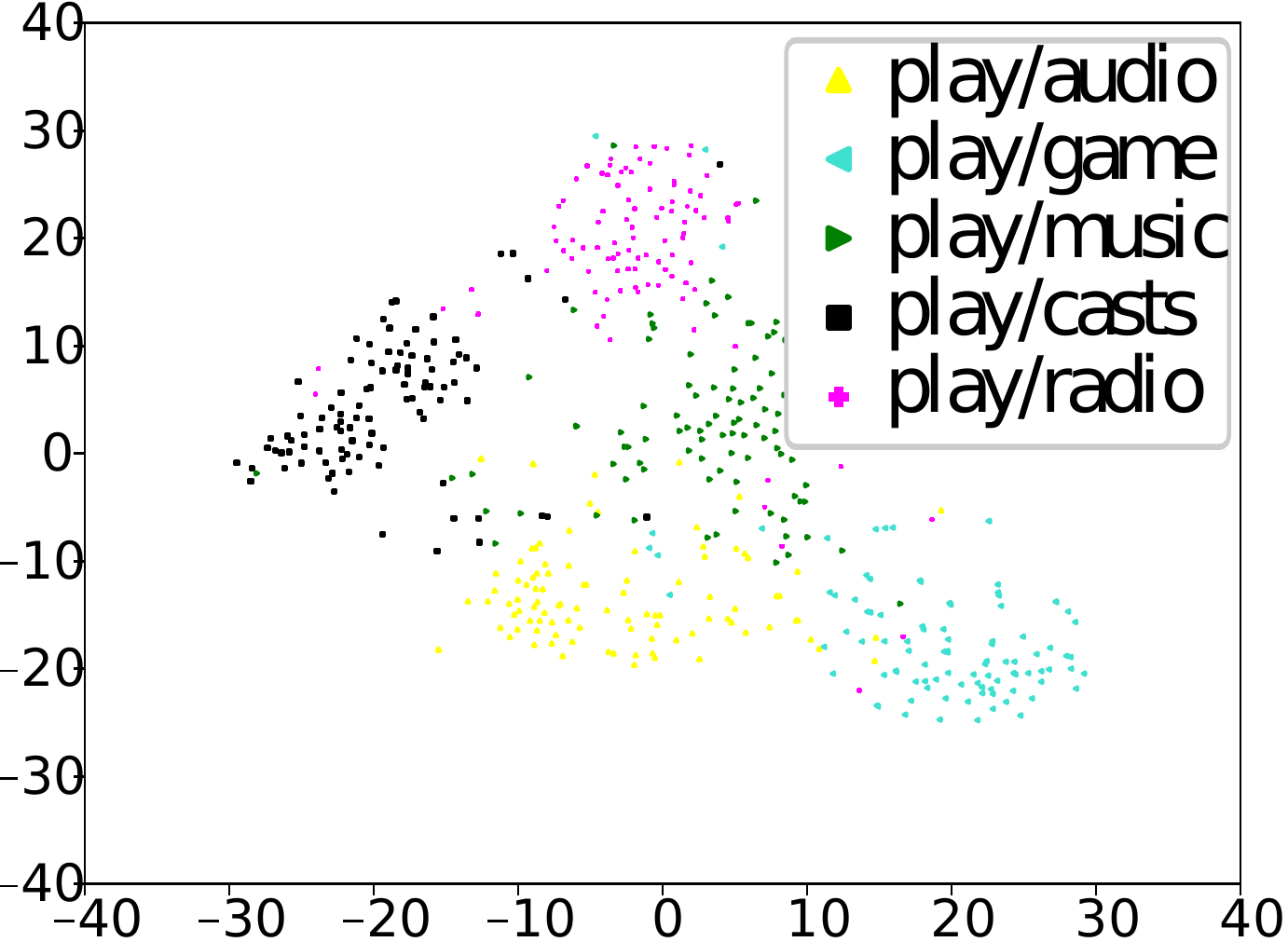}
			}\\
		\hspace{-.25cm}
		(a) Prototypical Networks & 
		\hspace{-.55cm}
		(b) ContrastNet
		\end{tabular}
	\end{center}
	\caption{Visualization of query text representations sampled from similar target classes on HWU64.}
	\label{fig:dis}
\end{figure}

To investigate models' ability in learning discriminative text representations of similar classes, we visualize the query-text representations produced by Prototypical Networks and ContrastNet using t-SNE~\cite{Maaten:08} in Figure~\ref{fig:dis}. We generate $100$ episodes in the 5-way 1-shot setting from the test set of HWU64, in which the text instances of query set are sampled from selected $5$ similar classes which all belong to the {\em play} domain and may provide texts with similar semantics. From Figure~\ref{fig:dis} (a), we observe that the text representations of similar classes produced by Prototypical Networks are prone to mix with each other, thus may make them hard to be distinguished by the prediction model. The text representations produced by ContrastNet in Figure~\ref{fig:dis} (b) are also not clearly separated, but they are much more discriminative than the query-text representations produced by Prototypical Networks. This visualization result demonstrates the power of ContrastNet in learning discriminative representations compared to Prototypical Networks.

\subsubsection{Error Analysis on Similar Classes}
\begin{figure}[ht]
	\begin{center}
		\begin{tabular}{cc}
		\hspace{-.3cm}
			\scalebox{0.32}{
		    \includegraphics{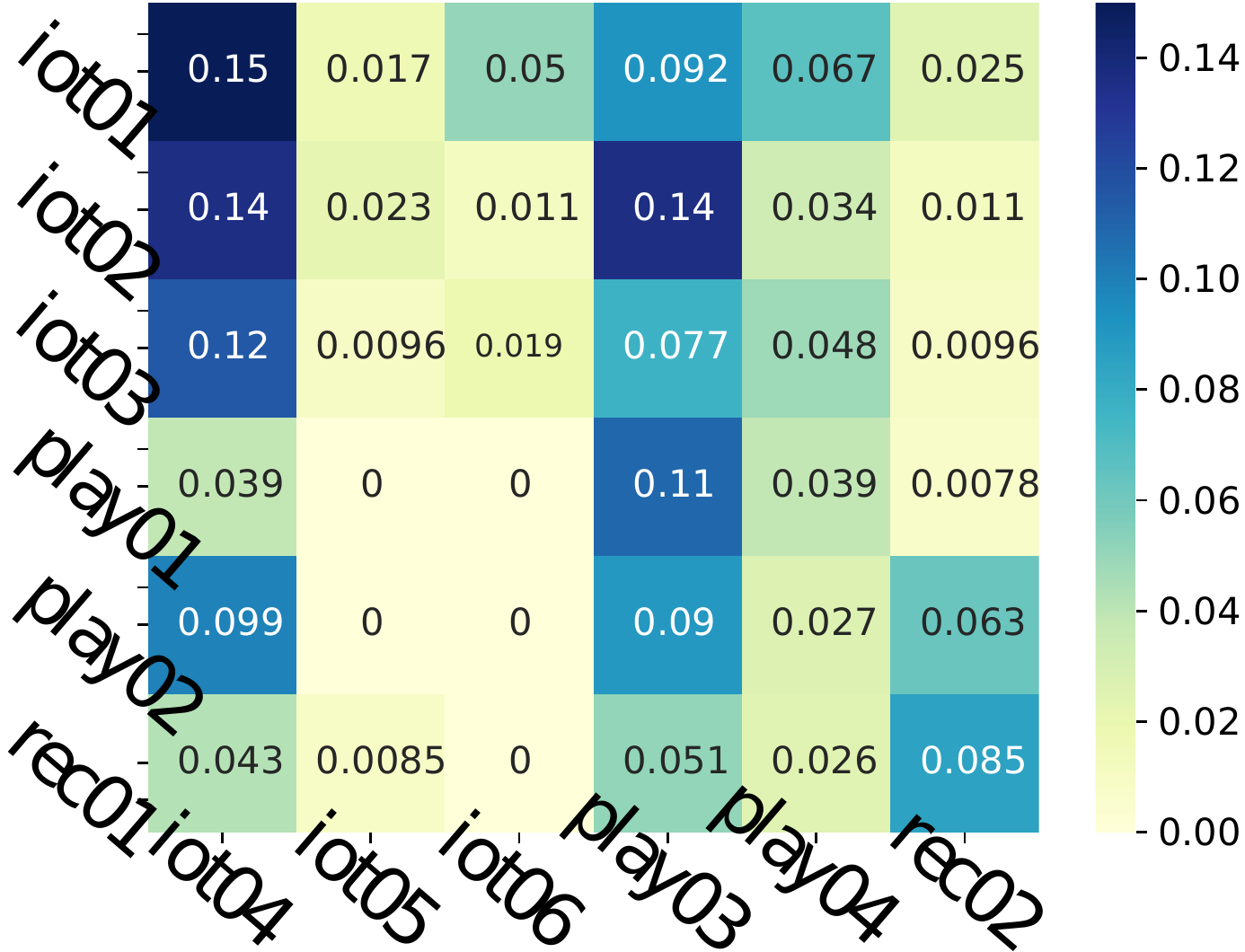}
			}&
		\hspace{-1.3cm}
			\scalebox{0.32}{
			\includegraphics{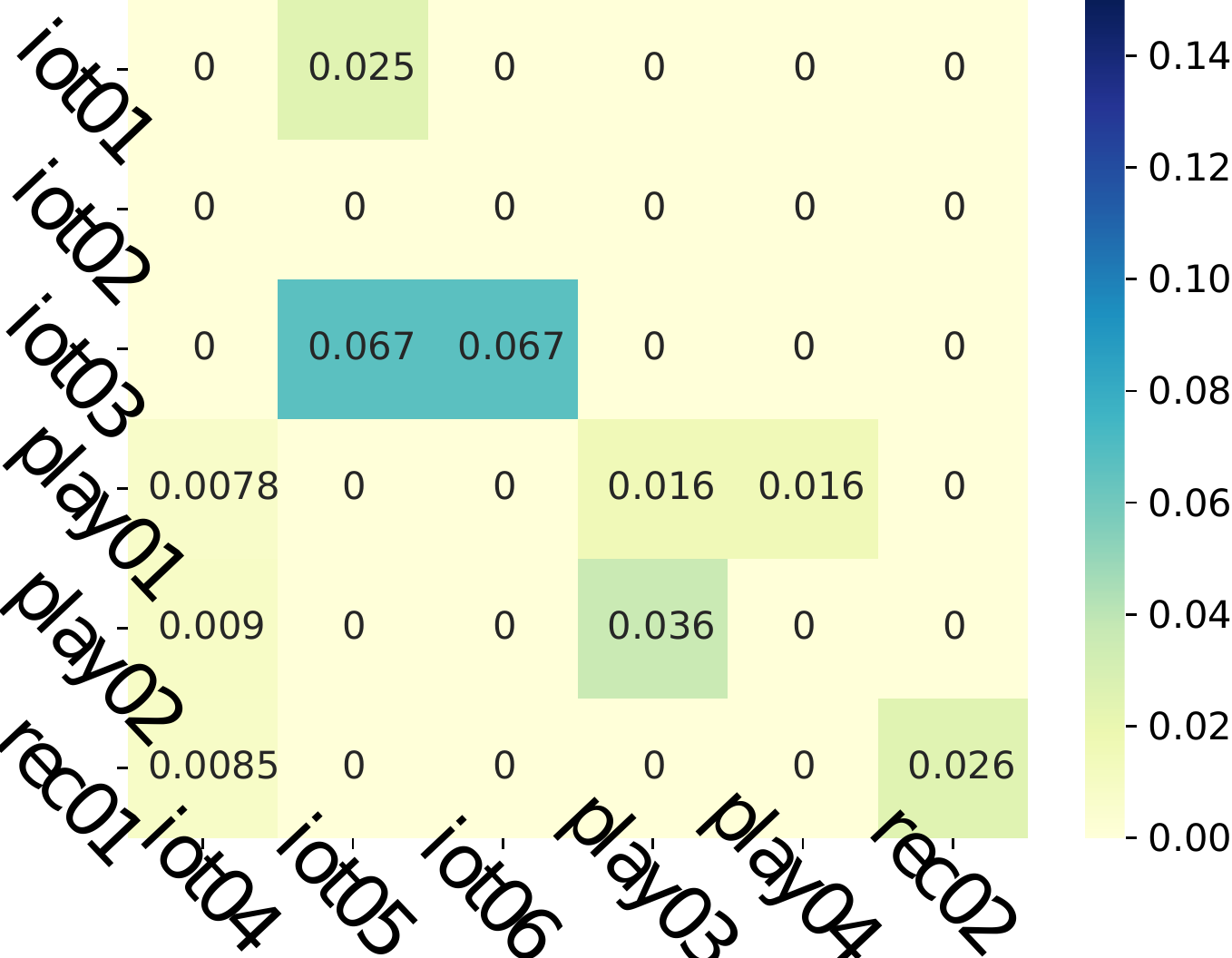}
			}\\
		\hspace{-.8cm}
		(a) Prototypical Networks & 
		\hspace{-1.0cm}
		(b) ContrastNet
		\end{tabular}
	\end{center}
	\caption{Error analysis of Prototypical Networks and ContrastNet on similar classes on HWU64. 
}
	\label{fig:error}
\end{figure}

To study whether improving the discrimination of text representations help improve few-shot text classification performance on similar classes, we make an error analysis of the prediction results on selected similar classes in the test set of HWU64. Each value in the heat-maps of Figure \ref{fig:error} denotes the proportion of query text instances of one class been misclassified to another class, e.g., Prototypical Networks misclassify $15$ percent of query text instances with class iot/lightoff ({\em iot01}) to class iot/coffee ({\em iot04}). Figure \ref{fig:error} (a) shows that the misclassification between similar classes is common  in the prediction results of Prototypical Networks. Figure \ref{fig:error} (b) shows ContrastNet significantly reduces the misclassification compared with Prototypical Networks. This observation suggests that by improving the discrimination of text representations, ContrastNet alleviates prediction contradictions between similar classes, thus improves the accuracy.

\subsection{Analysis of Unsupervised Regularization}
\subsubsection{Effectiveness of task-level Regularization}
\begin{figure}[ht]
	\begin{center}
		\begin{tabular}{cc}
		\hspace{-.25cm}
			\scalebox{0.29}{
		    \includegraphics{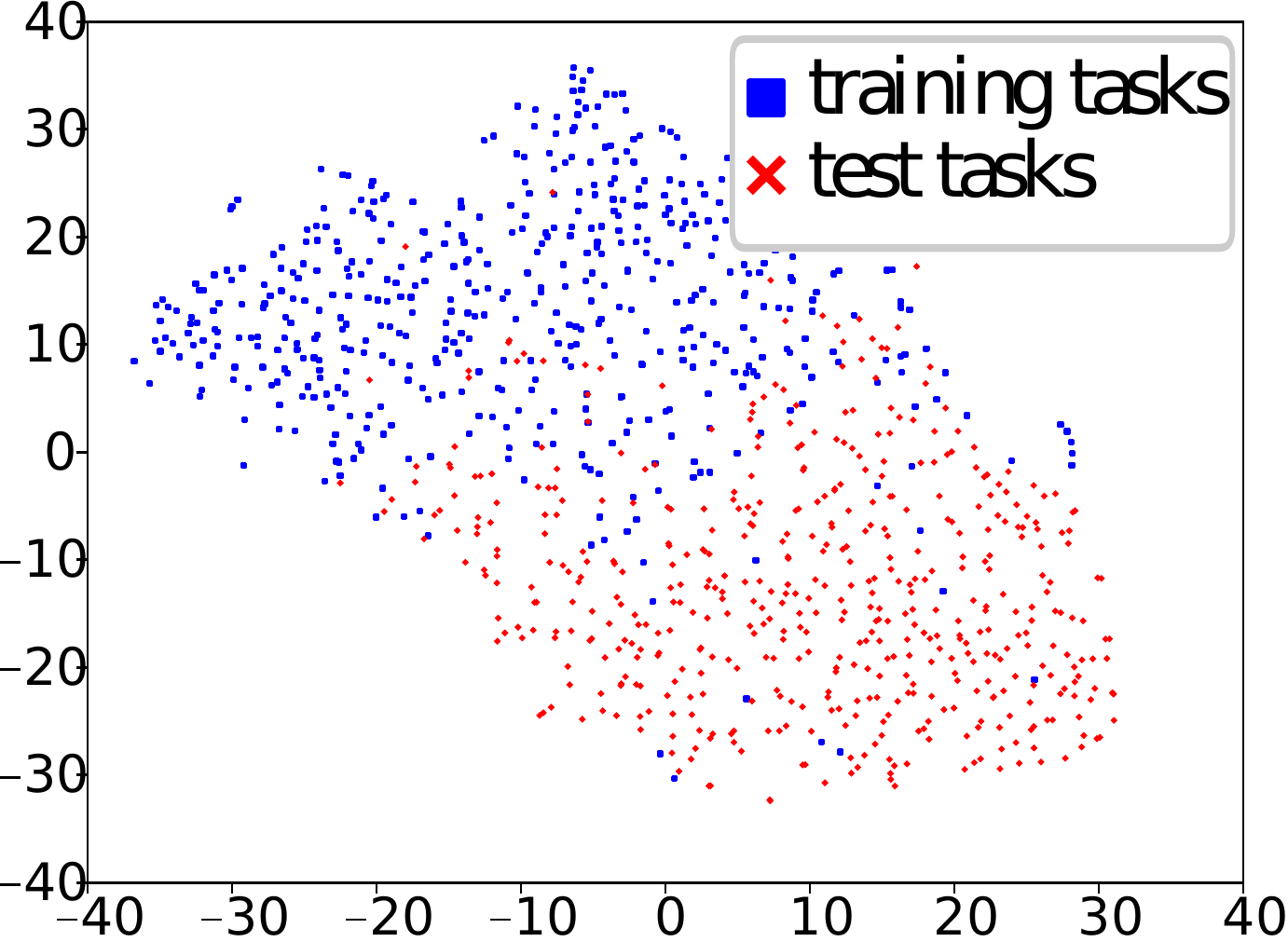}
			}&
		\hspace{-.57cm}
			\scalebox{0.29}{
			\includegraphics{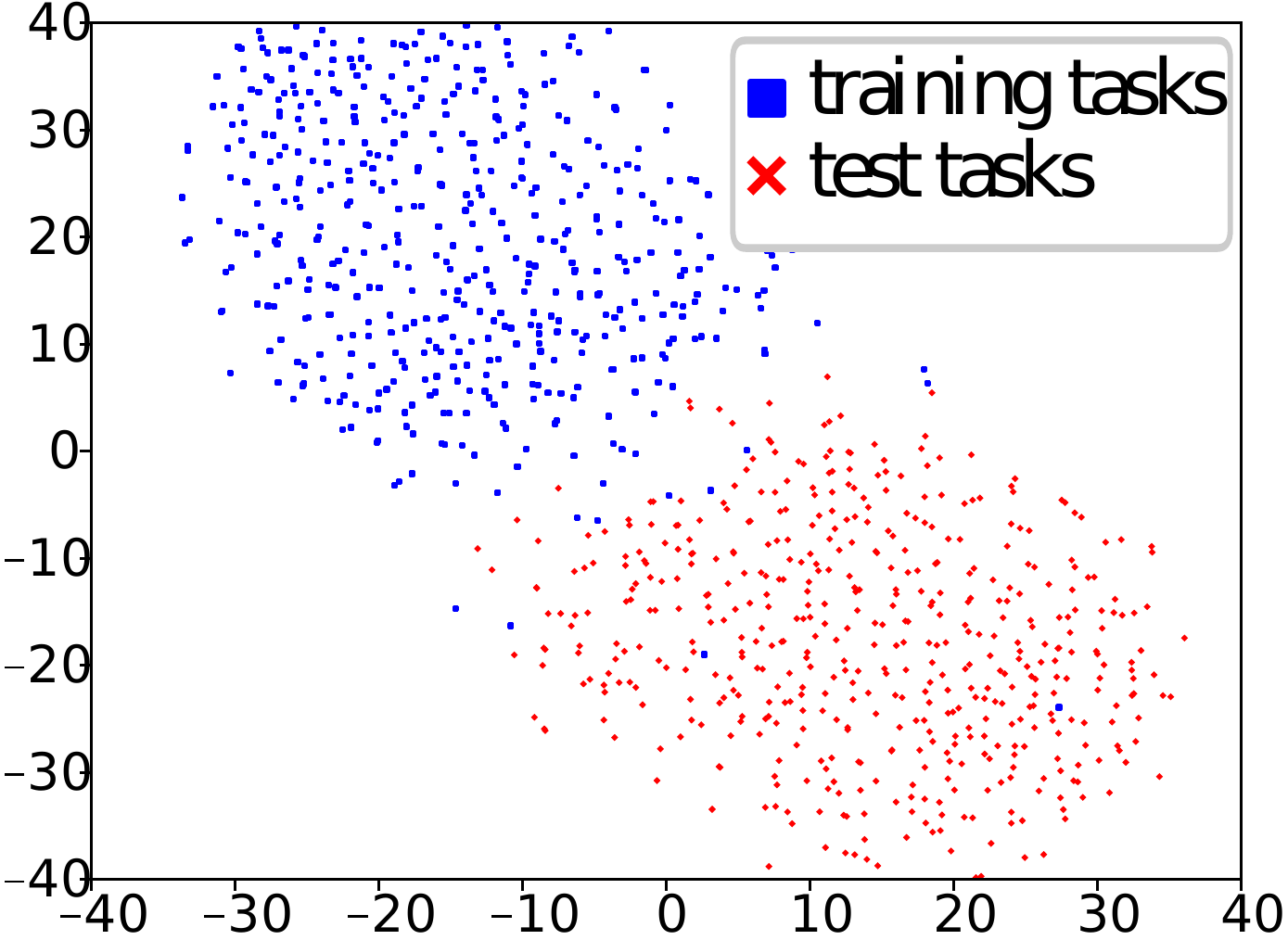}
			}\\
		\hspace{-.25cm}
		(a) Prototypical Networks & 
		\hspace{-.57cm}
		(b) ContrastNet
		\end{tabular}
	\end{center}
	\caption{Task-representation visualization of Prototypical Networks and ContrastNet on Banking77.}
	\label{fig:Reptask}
\end{figure}

To study whether ContrastNet learns more separable representations between training and testing tasks, we visualize the task representations on Banking77 using t-SNE. Specifically, We randomly sampled $200$ tasks from the training set and test set respectively and visualize the task representations produced by ContrastNet and Prototypical Networks in Figure \ref{fig:Reptask}. Figure \ref{fig:Reptask} (a) shows that the testing-task representations of Prototypical Networks are partially mixed with its training-task representations, i.e., overfit the training tasks. Figure \ref{fig:Reptask} (b) shows that the representations of training and testing tasks in ContrastNet are more separable than that in Prototypical Networks, which demonstrates the effectiveness of task-level regularization in ContrastNet.

\subsubsection{Effectiveness of Instance-level Regularization}
\begin{figure}[ht]
	\begin{center}
		\begin{tabular}{cc}
		\hspace{-.35cm}
			\scalebox{0.30}{
		    \includegraphics{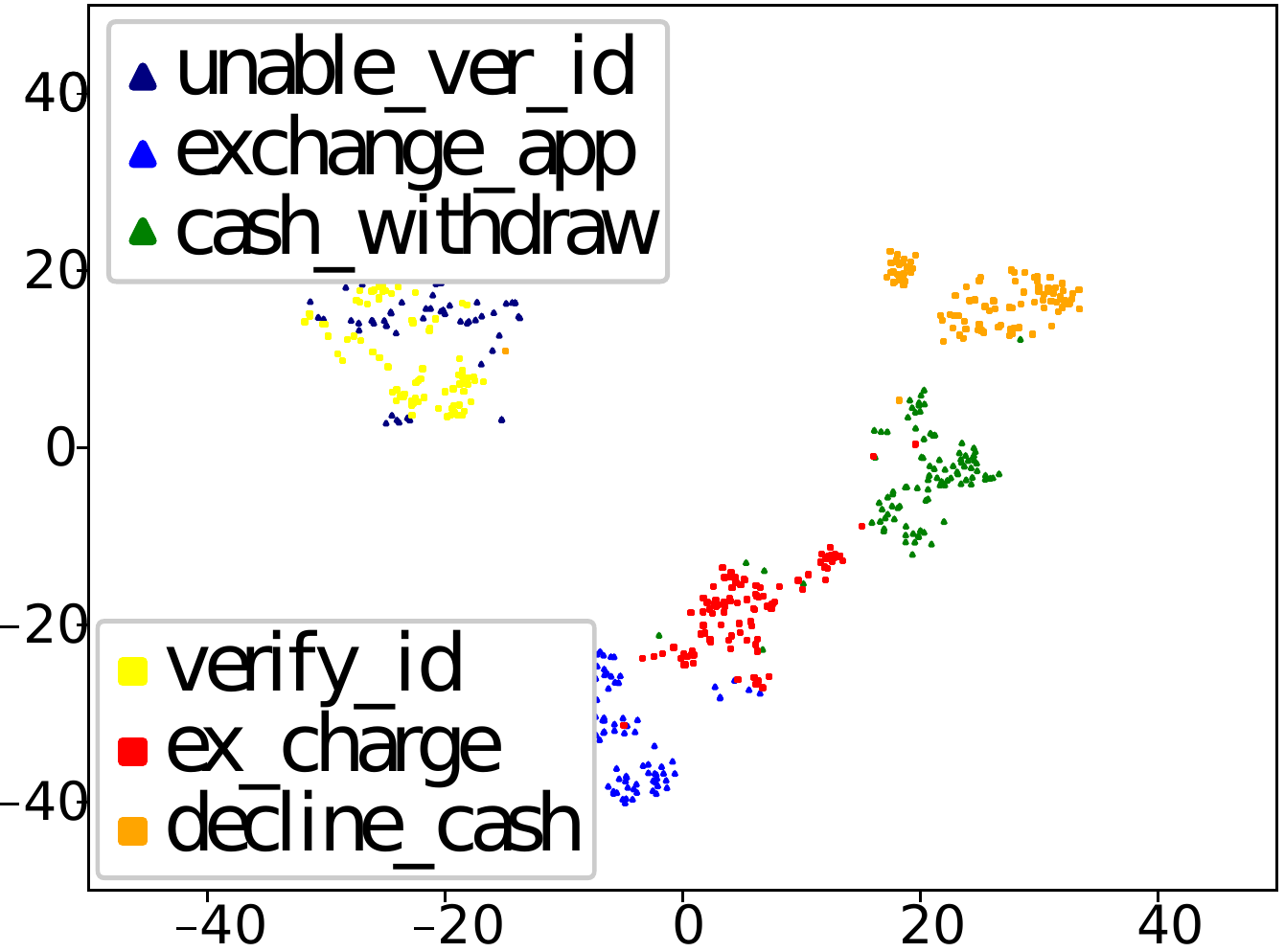}
			}&
		\hspace{-.5cm}
			\scalebox{0.30}{
			\includegraphics{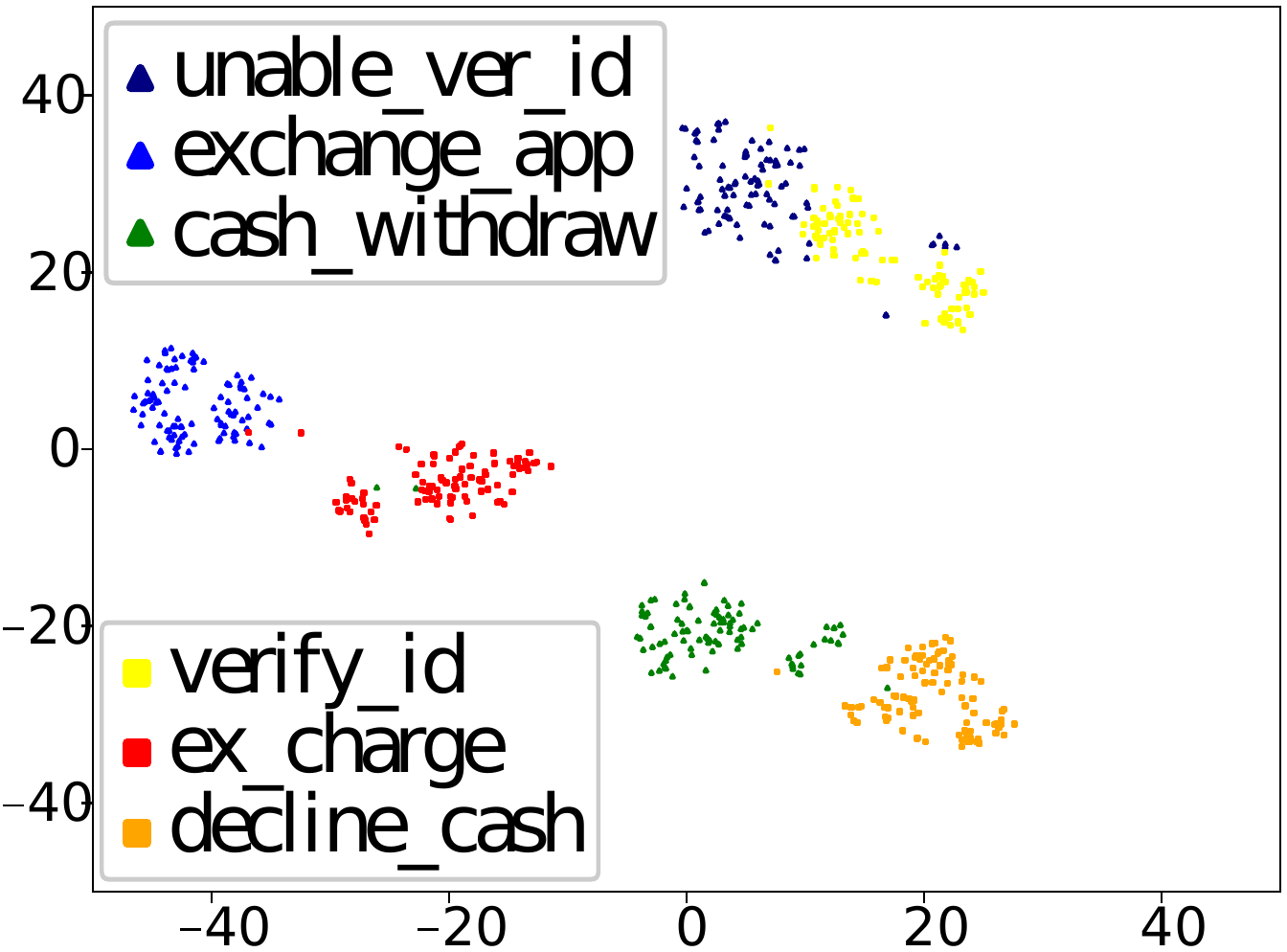}
			}\\
		\hspace{-.35cm}
		(a) Prototypical Networks & 
		\hspace{-.5cm}
		(b) ContrastNet
		\end{tabular}
	\end{center}
	\caption{Text-representation of Prototypical Network and ContrastNet on Banking77.}
	\label{fig:Repinst}
\end{figure}

We visualize the text representations of selected source and target classes to show whether the models learn separable text representations that alleviate the instance-level overfitting. Specifically, we select 3 source and target classes from the training and test set, respectively; and for each class, we randomly sample $100$ texts to visualize. As shown in Figure \ref{fig:Repinst} (a) and (b), the triangles with cool colors and squares with hot colors respectively denote source classes and target classes. Some text representations of target classes in Prototypical Networks locate near the text representations of source classes, i.e., overfit the training instances. In ContrastNet, the text representations of source and target classes are more separable from each other, which manifests the effectiveness of instance-level regularization in ContrastNet.

\section{Conclusion}
We propose a contrastive learning framework ContrastNet for few-shot text classification which learns discriminative text representation of similar classes and tackles the task and instance level overfitting problems. ContrastNet learns discriminative text representations belonging to different classes via supervised contrastive learning, while simultaneously introduce unsupervised contrastive regularization at both task and instance level to prevent overfitting. As the discriminative representation and overfitting problems are shared challenges in few-shot learning, we hope ContrastNet will extend to a broad spectrum of other applications.

\section*{Acknowledgments}
This work is supported partly by the National Natural Science Foundation of China (No. 61772059), by the Fundamental Research Funds for the Central Universities by the State Key Laboratory of Software Development Environment (No. SKLSDE-2020ZX-14).
\bibliography{aaai22}

\begin{thebibliography}{39}
\providecommand{\natexlab}[1]{#1}

\bibitem[{Bansal, Jha, and McCallum(2020)}]{Bansal:20}
Bansal, T.; Jha, R.; and McCallum, A. 2020.
\newblock Learning to Few-Shot Learn Across Diverse Natural Language
  Classification Tasks.
\newblock In \emph{COLING}, 5108--5123.

\bibitem[{Bao et~al.(2020)Bao, Wu, Chang, and Barzilay}]{Bao:20}
Bao, Y.; Wu, M.; Chang, S.; and Barzilay, R. 2020.
\newblock Few-shot Text Classification with Distributional Signatures.
\newblock In \emph{ICLR}.

\bibitem[{Cai et~al.(2020)Cai, Wang, Pan, Yao, and Mei}]{Cai:20}
Cai, Q.; Wang, Y.; Pan, Y.; Yao, T.; and Mei, T. 2020.
\newblock Joint Contrastive Learning with Infinite Possibilities.
\newblock In \emph{NeurIPS}.

\bibitem[{Casanueva et~al.(2020)Casanueva, Temcinas, Gerz, Henderson, and
  Vulic}]{Casanueva:20}
Casanueva, I.; Temcinas, T.; Gerz, D.; Henderson, M.; and Vulic, I. 2020.
\newblock Efficient Intent Detection with Dual Sentence Encoders.
\newblock \emph{CoRR}, abs/2003.04807.

\bibitem[{Chen and Zhang(2021)}]{Chen:21}
Chen, Q.; and Zhang, J. 2021.
\newblock Multi-Level Contrastive Learning for Few-Shot Problems.
\newblock \emph{CoRR}, abs/2107.07608.

\bibitem[{Chen et~al.(2020)Chen, Kornblith, Norouzi, and Hinton}]{Chen:20}
Chen, T.; Kornblith, S.; Norouzi, M.; and Hinton, G.~E. 2020.
\newblock A Simple Framework for Contrastive Learning of Visual
  Representations.
\newblock In \emph{ICML}, volume 119 of \emph{Proceedings of Machine Learning
  Research}, 1597--1607. {PMLR}.

\bibitem[{Dopierre, Gravier, and Logerais(2021)}]{Dopierre:21}
Dopierre, T.; Gravier, C.; and Logerais, W. 2021.
\newblock ProtAugment: Unsupervised diverse short-texts paraphrasing for intent
  detection meta-learning.
\newblock \emph{CoRR}, abs/2105.12995.

\bibitem[{Finn, Abbeel, and Levine(2017)}]{Finn:17}
Finn, C.; Abbeel, P.; and Levine, S. 2017.
\newblock Model-Agnostic Meta-Learning for Fast Adaptation of Deep Networks.
\newblock In \emph{ICML}, volume~70 of \emph{Proceedings of Machine Learning
  Research}, 1126--1135. {PMLR}.

\bibitem[{Gao et~al.(2019)Gao, Han, Liu, and Sun}]{Gao:19}
Gao, T.; Han, X.; Liu, Z.; and Sun, M. 2019.
\newblock Hybrid Attention-Based Prototypical Networks for Noisy Few-Shot
  Relation Classification.
\newblock In \emph{AAAI}, 6407--6414. {AAAI} Press.

\bibitem[{Gao, Yao, and Chen(2021)}]{GaoT:21}
Gao, T.; Yao, X.; and Chen, D. 2021.
\newblock SimCSE: Simple Contrastive Learning of Sentence Embeddings.
\newblock \emph{CoRR}, abs/2104.08821.

\bibitem[{Gao et~al.(2021)Gao, Fei, Liu, Lu, Xiang, and Huang}]{Gao:21}
Gao, Y.; Fei, N.; Liu, G.; Lu, Z.; Xiang, T.; and Huang, S. 2021.
\newblock Contrastive Prototype Learning with Augmented Embeddings for Few-Shot
  Learning.
\newblock \emph{CoRR}, abs/2101.09499.

\bibitem[{Geng et~al.(2020)Geng, Li, Li, Sun, and Zhu}]{Geng:20}
Geng, R.; Li, B.; Li, Y.; Sun, J.; and Zhu, X. 2020.
\newblock Dynamic Memory Induction Networks for Few-Shot Text Classification.
\newblock In \emph{ACL}, 1087--1094.

\bibitem[{Geng et~al.(2019)Geng, Li, Li, Zhu, Jian, and Sun}]{Geng:19}
Geng, R.; Li, B.; Li, Y.; Zhu, X.; Jian, P.; and Sun, J. 2019.
\newblock Induction Networks for Few-Shot Text Classification.
\newblock In \emph{EMNLP-IJCNLP}, 3902--3911. Association for Computational
  Linguistics.

\bibitem[{Han et~al.(2021)Han, Fan, Zhang, Qiu, Gao, and Zhou}]{Han:21}
Han, C.; Fan, Z.; Zhang, D.; Qiu, M.; Gao, M.; and Zhou, A. 2021.
\newblock Meta-Learning Adversarial Domain Adaptation Network for Few-Shot Text
  Classification.
\newblock In \emph{ACL/IJCNLP (Findings)}, 1664--1673. Association for
  Computational Linguistics.

\bibitem[{He and McAuley(2016)}]{He:16}
He, R.; and McAuley, J.~J. 2016.
\newblock Ups and Downs: Modeling the Visual Evolution of Fashion Trends with
  One-Class Collaborative Filtering.
\newblock In \emph{WWW}, 507--517. {ACM}.

\bibitem[{Hou et~al.(2019)Hou, Chang, Ma, Shan, and Chen}]{Hou:19}
Hou, R.; Chang, H.; Ma, B.; Shan, S.; and Chen, X. 2019.
\newblock Cross Attention Network for Few-shot Classification.
\newblock In \emph{NeurIPS}, 4005--4016.

\bibitem[{Kalantidis et~al.(2020)Kalantidis, Sariyildiz, Pion, Weinzaepfel, and
  Larlus}]{Kalantidis:20}
Kalantidis, Y.; Sariyildiz, M.~B.; Pion, N.; Weinzaepfel, P.; and Larlus, D.
  2020.
\newblock Hard Negative Mixing for Contrastive Learning.
\newblock In \emph{NeurIPS}.

\bibitem[{Khosla et~al.(2020)Khosla, Teterwak, Wang, Sarna, Tian, Isola,
  Maschinot, Liu, and Krishnan}]{Khosla:20}
Khosla, P.; Teterwak, P.; Wang, C.; Sarna, A.; Tian, Y.; Isola, P.; Maschinot,
  A.; Liu, C.; and Krishnan, D. 2020.
\newblock Supervised Contrastive Learning.
\newblock In \emph{NeurIPS}.

\bibitem[{Kingma and Ba(2015)}]{Kingma:15}
Kingma, D.~P.; and Ba, J. 2015.
\newblock Adam: {A} Method for Stochastic Optimization.
\newblock In \emph{{ICLR} 2015}.

\bibitem[{Lang(1995)}]{Lang:95}
Lang, K. 1995.
\newblock NewsWeeder: Learning to Filter Netnews.
\newblock In \emph{ICML}, 331--339. Morgan Kaufmann.

\bibitem[{Larson et~al.(2019)Larson, Mahendran, Peper, Clarke, Lee, Hill,
  Kummerfeld, Leach, Laurenzano, Tang, and Mars}]{Larson:19}
Larson, S.; Mahendran, A.; Peper, J.~J.; Clarke, C.; Lee, A.; Hill, P.;
  Kummerfeld, J.~K.; Leach, K.; Laurenzano, M.~A.; Tang, L.; and Mars, J. 2019.
\newblock EMNLP-IJCNLP.
\newblock 1311--1316. Association for Computational Linguistics.

\bibitem[{Liu et~al.(2021)Liu, Fu, Xu, Yang, Li, Wang, and Zhang}]{Liu:21}
Liu, C.; Fu, Y.; Xu, C.; Yang, S.; Li, J.; Wang, C.; and Zhang, L. 2021.
\newblock Learning a Few-shot Embedding Model with Contrastive Learning.
\newblock In \emph{AAAI}, 8635--8643. {AAAI} Press.

\bibitem[{Liu et~al.(2019{\natexlab{a}})Liu, Eshghi, Swietojanski, and
  Rieser}]{Liu:19}
Liu, X.; Eshghi, A.; Swietojanski, P.; and Rieser, V. 2019{\natexlab{a}}.
\newblock Benchmarking Natural Language Understanding Services for Building
  Conversational Agents.
\newblock In \emph{IWSDS}, volume 714 of \emph{Lecture Notes in Electrical
  Engineering}, 165--183. Springer.

\bibitem[{Liu et~al.(2019{\natexlab{b}})Liu, Eshghi, Swietojanski, and
  Rieser}]{LiuX:19}
Liu, X.; Eshghi, A.; Swietojanski, P.; and Rieser, V. 2019{\natexlab{b}}.
\newblock Benchmarking Natural Language Understanding Services for Building
  Conversational Agents.
\newblock In \emph{IWSDS}, volume 714 of \emph{Lecture Notes in Electrical
  Engineering}, 165--183. Springer.

\bibitem[{Luo et~al.(2021{\natexlab{a}})Luo, Liu, Lin, and Zhang}]{Luo:21}
Luo, Q.; Liu, L.; Lin, Y.; and Zhang, W. 2021{\natexlab{a}}.
\newblock Don't Miss the Labels: Label-semantic Augmented Meta-Learner for
  Few-Shot Text Classification.
\newblock In \emph{ACL/IJCNLP (Findings)}, 2773--2782. Association for
  Computational Linguistics.

\bibitem[{Luo et~al.(2021{\natexlab{b}})Luo, Chen, Wen, Pan, and Xu}]{LuoX:21}
Luo, X.; Chen, Y.; Wen, L.; Pan, L.; and Xu, Z. 2021{\natexlab{b}}.
\newblock Boosting Few-Shot Classification with View-Learnable Contrastive
  Learning.
\newblock \emph{CoRR}, abs/2107.09242.

\bibitem[{Majumder et~al.(2021)Majumder, Ravichandran, Maji, Polito, Bhotika,
  and Soatto}]{Majumder:21}
Majumder, O.; Ravichandran, A.; Maji, S.; Polito, M.; Bhotika, R.; and Soatto,
  S. 2021.
\newblock Revisiting Contrastive Learning for Few-Shot Classification.
\newblock \emph{CoRR}, abs/2101.11058.

\bibitem[{Mehri, Eric, and Hakkani{-}T{\"{u}}r(2020)}]{Mehri:20}
Mehri, S.; Eric, M.; and Hakkani{-}T{\"{u}}r, D. 2020.
\newblock DialoGLUE: {A} Natural Language Understanding Benchmark for
  Task-Oriented Dialogue.
\newblock \emph{CoRR}, abs/2009.13570.

\bibitem[{Snell, Swersky, and Zemel(2017)}]{Snell:17}
Snell, J.; Swersky, K.; and Zemel, R.~S. 2017.
\newblock Prototypical Networks for Few-shot Learning.
\newblock In \emph{NIPS}, 4077--4087.

\bibitem[{Sun et~al.(2021)Sun, Ouyang, Zhang, and Dai}]{Sun:21}
Sun, P.; Ouyang, Y.; Zhang, W.; and Dai, X. 2021.
\newblock {MEDA:} Meta-Learning with Data Augmentation for Few-Shot Text
  Classification.
\newblock In \emph{IJCAI}, 3929--3935. ijcai.org.

\bibitem[{Sung et~al.(2018)Sung, Yang, Zhang, Xiang, Torr, and
  Hospedales}]{Sung:18}
Sung, F.; Yang, Y.; Zhang, L.; Xiang, T.; Torr, P. H.~S.; and Hospedales, T.~M.
  2018.
\newblock Learning to Compare: Relation Network for Few-Shot Learning.
\newblock In \emph{2018 {IEEE} Conference on Computer Vision and Pattern
  Recognition, {CVPR} 2018, Salt Lake City, UT, USA, June 18-22, 2018},
  1199--1208. {IEEE} Computer Society.

\bibitem[{Tian et~al.(2020)Tian, Sun, Poole, Krishnan, Schmid, and
  Isola}]{Tian:20}
Tian, Y.; Sun, C.; Poole, B.; Krishnan, D.; Schmid, C.; and Isola, P. 2020.
\newblock What Makes for Good Views for Contrastive Learning?
\newblock In \emph{NeurIPS}.

\bibitem[{Tseng et~al.(2020)Tseng, Lee, Huang, and Yang}]{Tseng:20}
Tseng, H.; Lee, H.; Huang, J.; and Yang, M. 2020.
\newblock Cross-Domain Few-Shot Classification via Learned Feature-Wise
  Transformation.
\newblock In \emph{ICLR}.

\bibitem[{van~der Maaten and Hinton(2008)}]{Maaten:08}
van~der Maaten, L.; and Hinton, G. 2008.
\newblock Visualizing data using t-SNE.
\newblock \emph{Journal of machine learning research}, 9(Nov): 2579--2605.

\bibitem[{Vinyals et~al.(2016)Vinyals, Blundell, Lillicrap, Kavukcuoglu, and
  Wierstra}]{Vinyals:16}
Vinyals, O.; Blundell, C.; Lillicrap, T.; Kavukcuoglu, K.; and Wierstra, D.
  2016.
\newblock Matching Networks for One Shot Learning.
\newblock In \emph{NIPS}, 3630--3638.

\bibitem[{Wei and Zou(2019)}]{Wei:19}
Wei, J.~W.; and Zou, K. 2019.
\newblock {EDA:} Easy Data Augmentation Techniques for Boosting Performance on
  Text Classification Tasks.
\newblock In \emph{EMNLP-IJCNLP}, 6381--6387. Association for Computational
  Linguistics.

\bibitem[{Yang, Liu, and Xu(2021)}]{Yang:21}
Yang, S.; Liu, L.; and Xu, M. 2021.
\newblock Free Lunch for Few-shot Learning: Distribution Calibration.
\newblock In \emph{ICLR}. OpenReview.net.

\bibitem[{You et~al.(2020)You, Chen, Sui, Chen, Wang, and Shen}]{You:20}
You, Y.; Chen, T.; Sui, Y.; Chen, T.; Wang, Z.; and Shen, Y. 2020.
\newblock Graph Contrastive Learning with Augmentations.
\newblock In \emph{NeurIPS}.

\bibitem[{Yu et~al.(2018)Yu, Guo, Yi, Chang, Potdar, Cheng, Tesauro, Wang, and
  Zhou}]{Yu:18}
Yu, M.; Guo, X.; Yi, J.; Chang, S.; Potdar, S.; Cheng, Y.; Tesauro, G.; Wang,
  H.; and Zhou, B. 2018.
\newblock Diverse Few-Shot Text Classification with Multiple Metrics.
\newblock In \emph{NAACL-HLT}, 1206--1215. Association for Computational
  Linguistics.

\end{thebibliography}

%
%

\end{document}